\documentclass[10pt,twocolumn,letterpaper]{article}

\usepackage[pagenumbers]{cvpr} 

\newcommand{\red}[1]{{\color{red}#1}}

\definecolor{cvprblue}{rgb}{0.21,0.49,0.74}
\usepackage[pagebackref,breaklinks,colorlinks,allcolors=cvprblue]{hyperref}

\def\paperID{14431} 
\def\confName{CVPR}
\def\confYear{2026}

\title{F2IDiff: Real-world Image Super-resolution using Feature to Image Diffusion Foundation Model}

\author{Devendra
K. Jangid\ \ \
Ripon K. Saha\ \ \
Dilshan Godaliyadda\ \ \ 
Jing Li\ \
Seok-Jun Lee\ \   
Hamid R. Sheikh\\
 MPI Lab, Samsung Research America\\
 {\tt\small {\{d.jangid, dilshan.g, hr.sheikh\}}@samsung.com}
 }

\begin{document}
\maketitle
\begin{abstract}
With the advent of Generative AI, Single Image Super-Resolution (SISR) quality has seen substantial improvement, as the strong priors learned by Text-2-Image Diffusion (T2IDiff) Foundation Models (FM) can bridge the gap between High-Resolution (HR) and Low-Resolution (LR) images.
However, flagship smartphone cameras have been slow to adopt generative models because strong generation can lead to undesirable hallucinations.
For substantially degraded LR images, as seen in academia, strong generation is required and hallucinations are more tolerable because of the wide gap between LR and HR images. 
In contrast, for smartphone photography, the LR image has substantially higher fidelity, requiring only minimal hallucination-free generation.
We hypothesize that generation in SISR is controlled by the stringency and richness of the FM's conditioning feature.
First, text features are high level features, which often cannot describe subtle textures in an image. 
Additionally, Smartphone LR images are at least $12MP$, whereas SISR networks built on T2IDiff FM are designed to perform inference on much smaller images ($<1MP$). 
As a result, SISR inference has to be performed on small patches, which often cannot be accurately described by text feature. 
To address these shortcomings, we introduce an SISR network built on a FM with lower-level feature conditioning, specifically DINOv2 features, which we call a Feature-to-Image Diffusion (F2IDiff) Foundation Model (FM). 
Lower level features provide stricter conditioning while being rich descriptors of even small patches. 
We demonstrate the superiority of F2IDiff over T2IDiff FMs for SISR by training both with the same dataset and showing that F2IDiff achieves better fidelity through controlled generation.
Furthermore, Given the high fidelity of smartphone LR images and the richness of DINOv2 features, we demonstrate that the underlying FM can be trained with just 38K images (vs. billions in SD2.1), using a $2\times$ smaller U-Net, while achieving higher quality than SD2.1-based SISR.

\end{abstract}    
\section{Introduction}
\label{sec:intro}

\begin{figure}[t]
  \centering
   \includegraphics[width=1\linewidth]{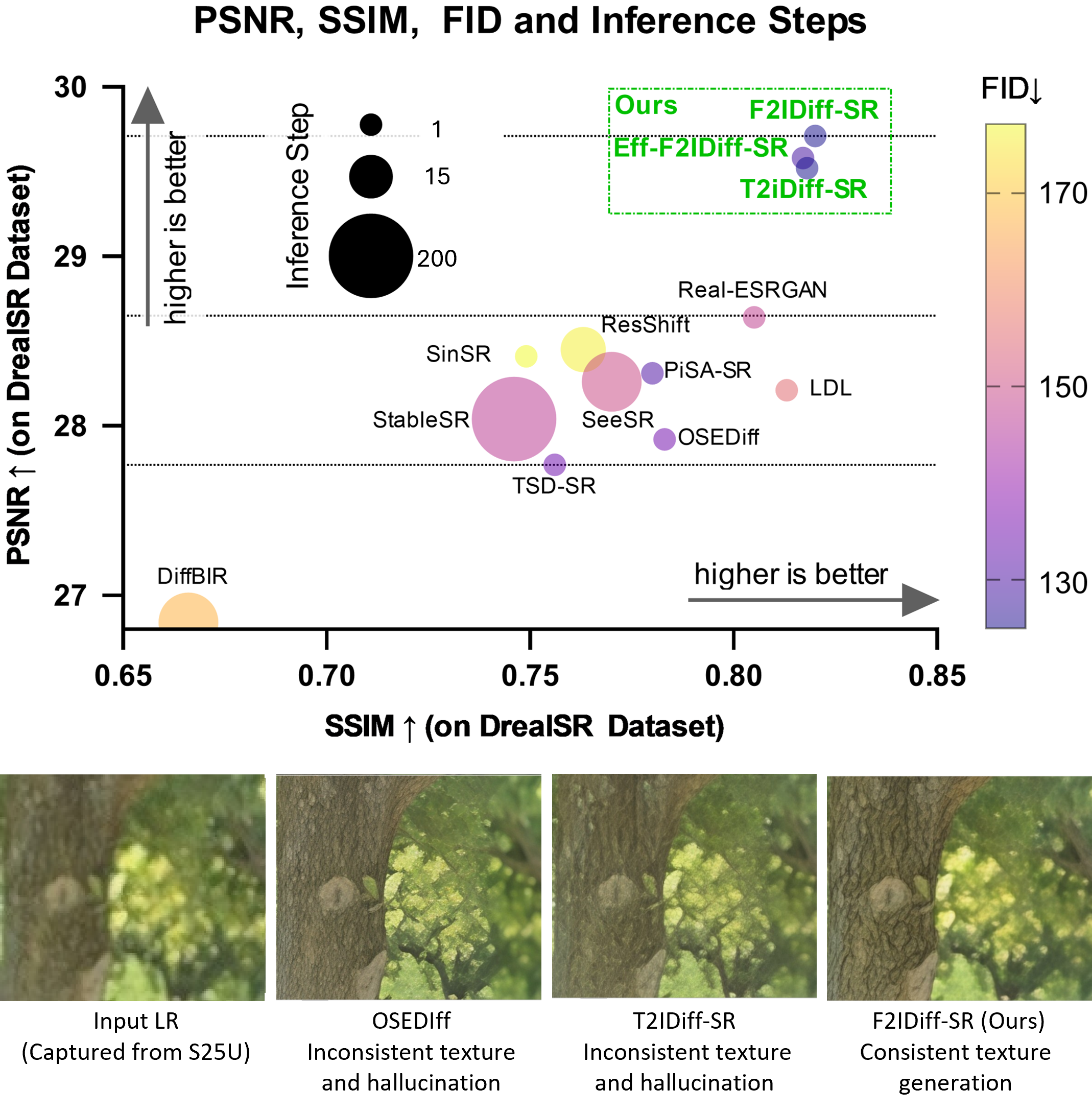}

   \caption{In top figure, F2IDiff-SR shows exceptional performance on the Real-SISR ($4\times$) task, achieving better metrics:- PSNR, SSIM, and FID. It outperforms SOTA methods by a substantial margin, highlighting its effectiveness. In the bottom figure, F2IDiff-SR gives better results compared to other methods on real-world images captured from an S25 Ultra smartphone. The other methods generate inconsistent texture and hallucinations.}
   \label{fig:PSNR_SSIM_FID}
\end{figure}
Single-image super-resolution (SISR)~\cite{SISR_wang2020deep} involves reconstructing a HR image from an degraded LR input. 
SISR is a classical yet active research problem with diverse applications, including smartphone cameras~\cite{ignatov2017dslr, ignatov2021real}, climate research~\cite{vandal2017deepsd, stengel2020adversarial}, material science~\cite{jangid2022adaptable, jangid2024q}, satellite imagery~\cite{shermeyer2019effects, nguyen2021self}, and medical science~\cite{georgescu2023multimodal, chen2023cunerf}. 
In this paper, we focus specifically on smartphone cameras. 
Most high-end smartphones come with multiple optical zoom lenses. 
The sensors mosaic patterns can enable $2\times$ and/or $4\times$ zoom, and optics allow for at most $10\times$ zoom. 
As a result, the remaining zoom scenarios have to be covered through digital zoom built on SISR.  Historically, classical image processing algorithms~\cite{keys2003cubic, tsai1984multiframe} were used to address this problem, but more recently, deep learning-based networks such as Convolutional Neural Networks (CNNs)~\cite{dong2015image, lim2017enhanced} and Transformers~\cite{liang2021swinir, lu2022transformer, chen2021pre} have been used more widely. 
Although early deep learning-based approaches significantly outperformed classical methods, they were largely discriminative rather than generative, and were commonly trained with maximum-likelihood objective implemented as pixel-wise L2/MSE. These methods tended to average over multimodal targets and produce overly smooth blurry outputs~\cite{ledig2017photo, johnson2016perceptual, blau2018perception}. 
To address this shortcoming, Generative Adversarial Network (GAN) based methods~\cite{Esrgan,RealEsrgan} employ adversarial loss objectives in an attempt to enhance the perceptual realism of the super-resolved images.  
However, such approaches can introduce artifacts and often compromise fidelity~\cite{ledig2017photo, FID}.  

\begin{figure}[t]
  \centering
   \includegraphics[width=1\linewidth]{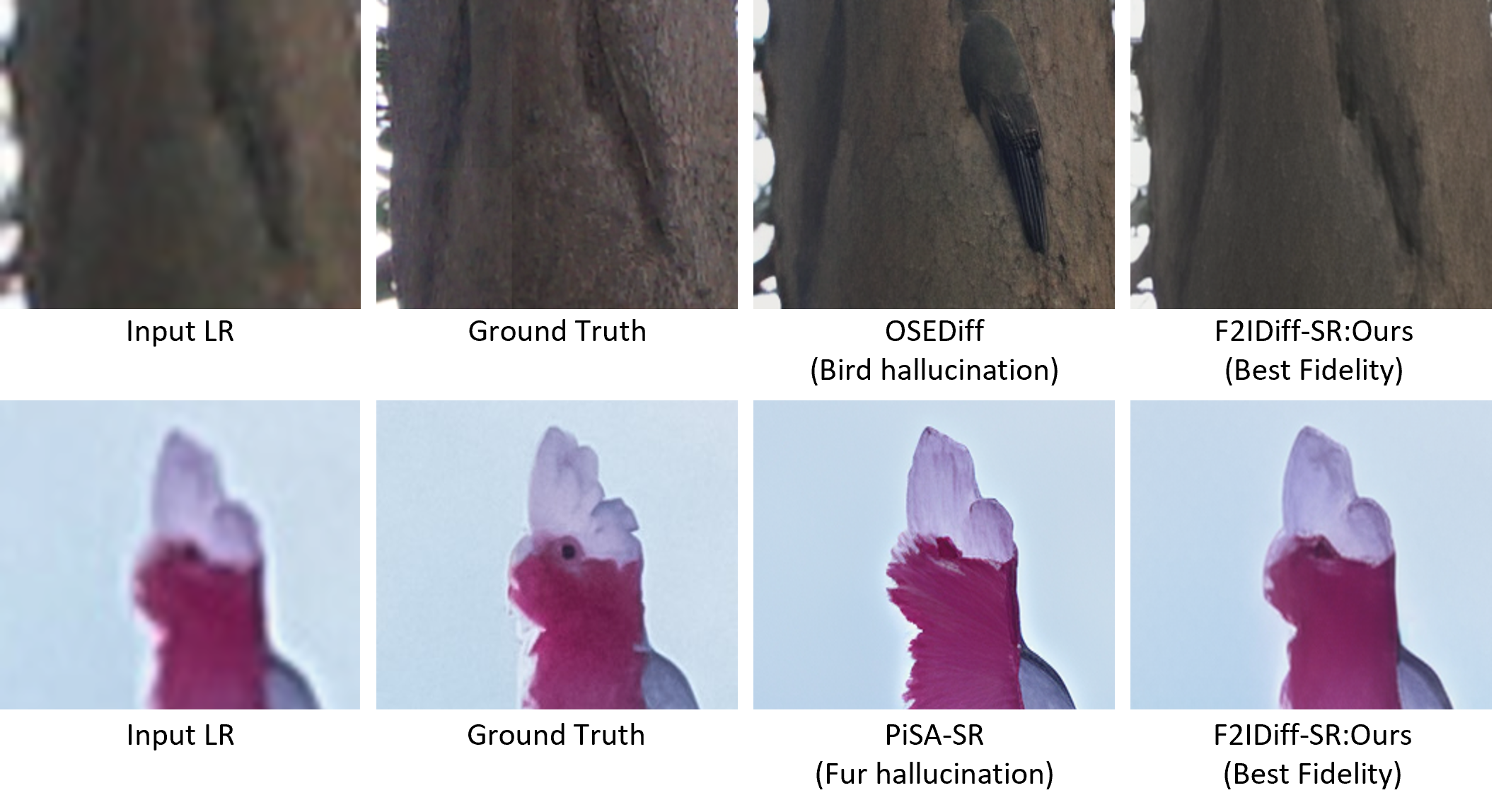}

   \caption{Zoom-in for details: SOTA single-step diffusion SR methods, such as OSEDiff~\cite{OSEDiff} and PiSA-SR~\cite{PiSASR}, produce excessive hallucination on real-world test datasets. For instance, OSEDiff generates a bird's tail in the output image even when no bird is present in the input. Similarly, PiSA-SR produces fur instead of facial features, leading to inaccurate and unrealistic results.}
   \label{fig:intro_hallucination}
\end{figure}

More recently, Diffusion-based methods have improved the generative capability of SISR substantially~\cite{OSEDiff,PiSASR,TSDSR,li2025diffusion}. These methods can be broadly categorized based on their underlying architectures.
One prominent family of approaches are based on pre-trained text-to-image diffusion (T2IDiff) FMs~\cite{PiSASR, OSEDiff, TSDSR}, such as Stable Diffusion (SD)~\cite{Rombach_2022_CVPR_SD2.1} which provides a strong prior to sample from. 
However, these methods are often complex, compute/memory intensive, and tend to produce significant hallucination artifacts as shown in Figure \ref{fig:PSNR_SSIM_FID} and \ref{fig:intro_hallucination}. 
While distillation techniques~\cite{fsalimans2022progressive_distillation1, song2023consistency_distillation2, noroozi2024you_distillation3, adcsr, Sinsr, TSDSR} can alleviate some of these complexity issues, they frequently result in a notable degradation in output quality~\cite{adcsr}.
The other family of approaches involve training diffusion models from scratch~\cite{Rombach_2022_CVPR_SD2.1, nichol2021glide_trainFromScratch1, saharia2022photorealistic, chen2023pixart} without relying on pre-trained FMs. 
These methods typically employ multi-step inference processes but often fall short in terms of generative quality, making them less appealing for real-world applications.
We posit that hallucinations in FM based methods are correlated to the rigidity and richness of the FM conditioning. 
In academia, most LR images used for SISR are severely degraded compared to their corresponding HR image. 
Therefore, hallucinations in SISR are useful, due to the large LR-HR gap. 
However, in smartphones, especially flagship devices, LR images have very high fidelity because they use high-end sensors and optics. 
For example Galaxy S25 Ultra ($1/1.3^{\prime\prime}$, 200MP, $f/1.7$), iPhone 17 Pro/Max ($\approx1/1.28^{\prime\prime}$, 48MP, $f/1.78$) and Google Pixel 10 Pro ($\approx1/1.3^{\prime\prime}$, 50MP, $f1/1.85$) - yielding superior SNR, dynamic range, and fine-texture details~\cite{SamsungGalaxyS25UltraSpecsPhoneArena, GooglePixel10ProSpecsPhoneArena, iPhone17ProMaxSpecsPhoneArena}. 
Our target domain is consumer photos, where preserving scene fidelity matters more than adding plausible texture. We therefore prefer controlled, hallucination-free enhancement over the aggressive hallucinations seen in some academic settings. Furthermore, in smartphones, the LR image is minimally $12MP$ i.e. $4k \times 3k$ while T2IDiff FMs~\cite{Rombach_2022_CVPR_SD2.1, ramesh2021zero_FM1,  ramesh2022hierarchica_FM2, saharia2022photorealistic} can only operate on at most $1k \times 1k$ images.
As a result, the image needs to be broken down into patches and super-resolved independently.
This means that the text caption must be generated for a patch of a higher resolution image, which often lacks sufficient content/context for a text captioning engine to generate a meaningful caption. 
For example, a patch may contain a portion of a face or a tree, and the captioning engine cannot meaningfully describe it, leading to erroneous conditioning. 
In fact the authors of OSEDiff have observed that giving a null vector as text conditioning often generated comparable results to text conditioning~\cite{OSEDiff}. 

To overcome these shortcomings, we propose a SISR network built on low-level feature-to-image Diffusion FMs (F2IDiff FM) instead of T2IDiff FM. 
In particular, we propose using DINOv2 features~\cite{oquab2023dinov2}, as it can capture low-levels details such as texture, whereas text tends to capture high-level semantic information. 
Furthermore, DINOv2 can capture differentiating features even at the patch level since it works at much lower level compared to text~\cite{jiang2023clip, barsellotti2025talking}.
Given that we only need to enhance an already high-fidelity input, we further posit that the underlying FM model can be trained with significantly fewer image, as long as they are carefully chosen.
In this paper, we use just 38,000 HR images to train our F2IDiff FM, compared to the billions used by SD 2.1~\cite{Rombach_2022_CVPR_SD2.1}, which is the basis of most SISR networks that use diffusion-based FMs.
Lastly, since we are using significantly fewer images to build a prior, the model capacity can be reduced significantly and we show comparable results with $2\times$ reduction in U-Net complexity. In summary, our contributions are:-
\begin{itemize}
    \item We introduce a Feature-to-Image Diffusion (F2IDiff) FM specifically tailored for the SISR problem, where the FM is conditioned on image features as opposed to text. 
    \item We develop an SISR network on F2IDiff FM, demonstrating superior performance compared to T2IDiff FM-based networks and state-of-the-art (SOTA) methods~\cite{PiSASR, OSEDiff, TSDSR} on both public datasets and real images captured using the Samsung S25U device. 
    \item We show that for SISR, a FM trained on just 38K carefully selected HR images, leveraging richer DINOv2 features instead of text, significantly outperforms models trained on billions of images, achieving superior fidelity and less hallucination. Additionally, this approach allows for a two-fold reduction in U-Net complexity. 
\end{itemize}

\section{Related Work}
\label{sec:related_work}
The field of SISR has advanced significantly, driven by deep learning methods. Here, we present the progression from early CNN approaches to modern generative models, positioning our F2IDiff-SR within the latest SOTA methods.

\textbf{GAN-based Real-world SISR:} Early deep learning methods for SISR were significantly advanced by GAN~\cite{goodfellow2014generativeadversarialnetworks}, which moved beyond pixel-wise metrics to achieve photorealistic results. SRGAN~\cite{saharia2022photorealistic}, its successor ESRGAN~\cite{Esrgan} and LDL~\cite{LDL_liang2022details} were seminal in this area. A critical challenge was their reliance on simple, known degradations. To improve generalization to real-world images, BSRGAN~\cite{zhang2021designing} and Real-ESRGAN~\cite{RealEsrgan} introduced high-order degradation modeling to synthesize more realistic training data. Despite their success in generating sharp details, GAN-based methods are known for training instability and a tendency to produce visual artifacts~\cite{arjovsky2017wassersteingan}, a limitation that persists in subsequent works.

\textbf{Multi-step Diffusion Models for Real-world SISR:} More recently, diffusion models showed better quality for SISR, leveraging powerful priors from pre-trained T2IDiff models like SD~\cite{Rombach_2022_CVPR_SD2.1}. A common paradigm is to guide the iterative denoising process with the LQ input, using techniques like fine-tuned encoders in StableSR~\cite{StableSR}, pre-restoration modules in DiffBIR~\cite{Diffbir}. To better incorporate semantic information, methods like SeeSR~\cite{Seesr} and SUPIR~\cite{yu2024scaling} utilize text prompts derived from the image content. Seeking richer context, some works have expanded conditioning beyond text to fuse multiple modalities, such as depth and segmentation, to improve detail recovery and reduce hallucinations~\cite{mei2025powercontextmultimodalityimproves}. While these approaches achieve exceptional perceptual quality, their iterative nature requires substantial computational resources and inference time, limiting their practical applicability~\cite{Resshift, Sinsr}.

\textbf{One-Step Diffusion for Real-world SISR:} To address the latency of multi-step methods, SinSR~\cite{Sinsr} has focused on distilling the generative capabilities of large diffusion models into a single forward pass. This is often enabled by techniques like Variational Score Distillation (VSD)~\cite{wang2023prolificdreamer_VSDLoss}. OSEDiff~\cite{OSEDiff} is a prominent example, using the LQ image as the starting point and a VSD loss to regularize the output. However, recent works have identified limitations in existing one-step models, such as unsatisfactory detail recovery or the generation of visual artifacts~\cite{TSDSR}. Consequently, new distillation frameworks have been proposed, including Target Score Distillation (TSD-SR)~\cite{TSDSR} which uses real image references, and Consistent Score Identity Distillation~\cite{wang2025gendrlightninggenerativerestorator} which tailors the VAE architecture with a larger latent space better suited for SR. Another critical research direction is providing user control on SISR task, and methods like PiSA-SR~\cite{PiSASR} and RCOD~\cite{wu2025realismcontrolonestepdiffusion} aim to resolve the inherent trade-off between fidelity and realism by allowing for adjustable control at inference time. PiSA-SR~\cite{PiSASR} achieves this by decoupling pixel and semantic objectives into two Low-Rank Adaptation (LoRA) modules, while RCOD~\cite{wu2025realismcontrolonestepdiffusion} employs a latent domain grouping strategy. Different from all of these methods, our F2IDiff-SR is built on our F2IDiff FM instead of a public T2IDiff FM. 


\begin{figure*}
  \centering
    \includegraphics[width=1\linewidth]{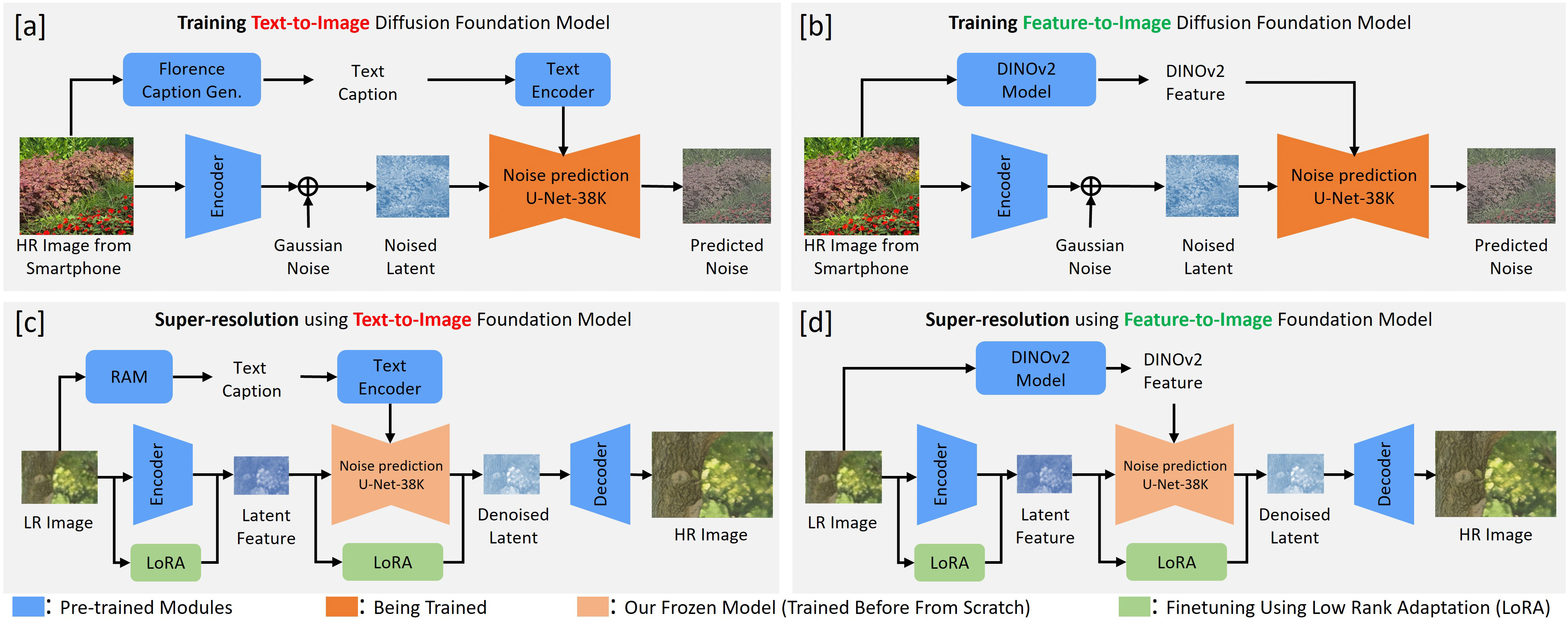}
    \caption{Our Methods:(a) Training pipeline of T2IDiff FM: A diffusion U-Net with text conditioning is trained from scratch on internal 38K HR images using a pre-trained Encoder-Decoder, Florence caption generation, and text-encoder. (b) Training pipeline of F2IDiff FM: A diffusion U-Net using DINOv2 features as conditioning is trained from scratch on internal 38K HR images using a pre-trained Encoder-Decoder, DINOv2 feature extractor. (c) SISR network based on T2IDiff FM: A single-step diffusion SR model is built on T2IDiff FM using LoRA. (d) SISR network based on F2IDiff FM: A single-step diffusion SR model is built on F2IDiff FM using LoRA.}
    \label{fig:arch_diagram}
    
\end{figure*}
\section{Methodology}
\label{sec:method}
In this section, we first elaborate on the training method of the F2IDiff FM. Then, we discuss how we integrate the trained F2IDiff FM model to a SISR network for $4\times$ super-resolution, utilizing the LoRA training strategy~\cite{hu2022lora}. Finally, we outline dataset collection and preparation that was employed for training both the FMs and the SISR network.
 

\subsection{Feature-to-Image Foundation Model}
We trained and developed both the F2IDiff and T2IDiff FMs on an internal dataset of 38K HR images to ensure a fair comparison between F2IDiff-SR and T2IDiff-SR for the SISR task. The objective was to demonstrate that F2IDiff-SR produces images with better fidelity and controlled generation, resulting in more realistic outputs compared to T2IDiff-SR. Both FMs' U-Nets were trained from scratch, as illustrated in Figure \ref{fig:arch_diagram} (a) and (b). For F2IDiff FM, DINOv2 features were used as low-level image conditioning due to their robustness and conciseness compared to text captions, while T2IDiff FM employed a pre-trained Clip ViT-L text encoder~\cite{radford2021learning} for feature extraction. Both FMs utilized a pre-trained encoder-decoder (f8d4)~\cite{Rombach_2022_CVPR_SD2.1} and were trained on approximately 1.5 million patches generated from 38K HR images captured by flagship smartphones. Given the informativeness of DINOv2 features and the limited dataset size (38K images), the U-Net complexity for F2IDiff FM was reduced twofold, involving a decrease in channels and attention head dimensions, resulting in Eff-F2IDiff, which outperforms SOTA methods for SISR. This approach highlights the superiority of F2IDiff-SR in generating realistic images for real-world SISR tasks.

\subsection{Single Image Super-resolution Network}
In this subsection, we design SISR networks using T2IDiff and F2IDiff FMs. 
Our objective is to develop an advanced SISR network for smartphone cameras that preserves fidelity and exhibits controlled generative capabilities, minimizing hallucinations and artifacts. 
To achieve this, we employ the LoRA strategy~\cite{OSEDiff, hu2022lora} to train our SISR networks using FMs:T2IDiff, F2IDiff, and Eff-F2IDiff, which were initially trained from scratch on our 38K HR dataset. 
We demonstrate empirically that the SISR network built on the F2IDiff FM significantly outperforms the SISR network built on the T2IDiff FM. 
For SISR network training, we generate LR images from HR images using the Real-ESRGAN~\cite{RealEsrgan} synthetic degradation pipeline to ensure a fair comparison with existing SOTA methods~\cite{PiSASR, OSEDiff, Sinsr}. 
However, we observe that in the context of SISR-based digital zoom for smartphones, the images captured by smartphone sensor have higher resolution and more details than the synthetic LR images, generated by synthetic degradation pipeline from public datasets.
To train the SISR network using the T2IDiff FM, we adopt the pipeline described in OSEDiff~\cite{OSEDiff}.
We replace the Stable Diffusion U-Net in this pipeline with our custom T2IDiff U-Net, which has been trained on 38K internal HR images, as opposed to the SDV2.1 U-Net trained on billions of public images. 
For the SISR network built on F2IDiff, we extract DINOv2 features from LR images, assuming that the difference between DINOv2 features of LR and HR images is minimal. 
These extracted DINOv2 features serve as conditioning for our F2IDiff U-Net, and we employ the LoRA strategy~\cite{hu2022lora} to train LoRA layers for both the U-Net and the pre-trained encoder. 
To compute the VSD loss~\cite{wang2023prolificdreamer_VSDLoss}, we use DINOv2 features instead of text features for our F2IDiff FM. Additionally, we train an efficient SISR network based on Eff-F2IDiff and compare its results against SOTA SR methods.  
\begin{figure*}
  \centering
    \includegraphics[width=1\linewidth]{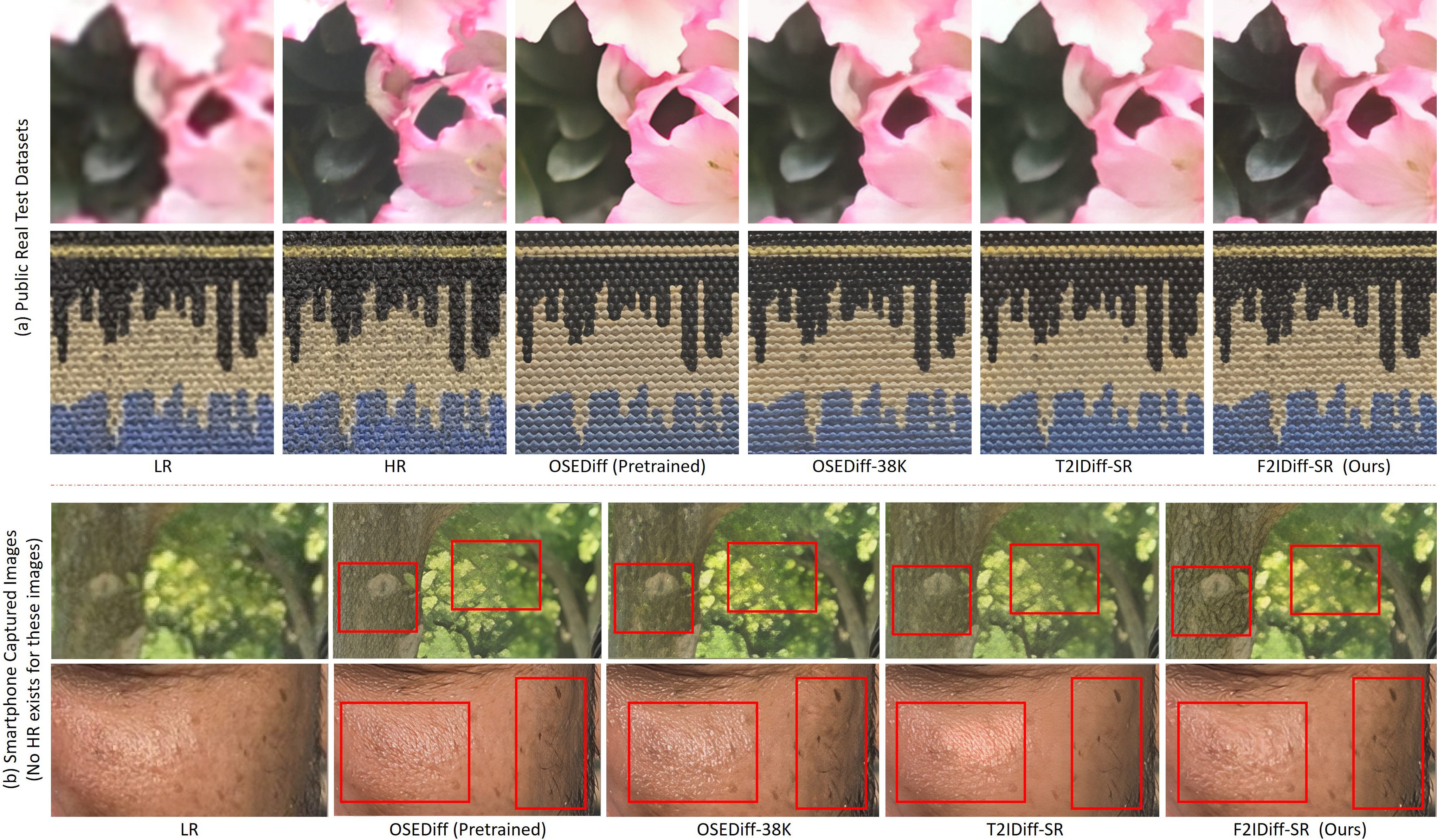}
    \caption{Zoom-in for best visuals: Qualitative comparison between our methods on public datasets and smartphone captured images. (a) Our method on public datasets show the highest fidelity with HR, while OSEDiff~\cite{OSEDiff}, which uses public SDV2.1~\cite{Rombach_2022_CVPR_SD2.1}, generates artificial textures. (b) Our method generates uniform textures and natural textures compared to other methods which use public SDV2.1~\cite{Rombach_2022_CVPR_SD2.1}.}
    \label{fig:qual_results_1}
    
\end{figure*}

\begin{table*}
  \caption{Quantitative comparison between T2IDiff-SR and F2IDiff-SR for $4\times$ SISR. For reference, we include pre-trained OSEDiff~\cite{OSEDiff}, and OSEDiff model trained on our 38K HR datasets (OSEDiff-38K). The best results are highlighted in \textbf{\color{red}{red}}. Our method F2IDiff-SR demonstrates superior performance on reference-based metrics on real-world datasets such as RealSR~\cite{RealSR} and DRealSR~\cite{DrealSR}. This is achieved despite our F2IDiff FM being trained on only 38K HR images, in contrast to the Public-T2IDiff FM model (SDV2.1)~\cite{Rombach_2022_CVPR_SD2.1}, which was trained on billion of images. On synthetic datasets like DIV2K~\cite{DIV2k}, which are less representative of real-world scenarios encountered in smartphone cameras due to their HR nature, our methods excel in fidelity metrics such as PSNR and SSIM.}
  \label{tab:quant_table_1}
  \centering
  \scriptsize
  \begin{tabular}{p{0.8cm}|p{1.5cm}|p{1.3cm}|p{0.9cm} p{0.8cm} p{0.9cm} p{1cm} p{0.7cm}|p{0.8cm} p{1.2cm} p{1cm} p{1.3cm}}
    \hline
    Data & Models & Base Method & \multicolumn{5}{c|}{Reference-based IQA - Focus on Fidelity via HR Comparison} & \multicolumn{4}{c}{Blind IQA - Focus on Perception without looking at HR} \\
    \cline{4-12}
    & & & PSNR $\uparrow$ & SSIM $\uparrow$ & LPIPS $\downarrow$ & DISTS $\downarrow$ & FID $\downarrow$ & NIQE $\downarrow$ & CLIPIQA $\uparrow$ & MUSIQ $\uparrow$ & MANIQA $\uparrow$ \\
    \hline

    \multirow{3}{4em}{DRealSR (Real Datasets)} 
    & OSEDiff~\cite{OSEDiff} &  Pub-T2I-FM
    & 27.92 & 0.783 & 0.297 & 0.217 & 135.29 & \textbf{\color{red}{6.49}} & \textbf{\color{red}{0.696}} & \textbf{\color{red}{64.65}} & \textbf{\color{red}{0.590}} 
    \\ 
    & OSEDiff-38K &  Pub-T2I-FM
    & 29.51 & 0.816 & 0.249 & 0.199 & 133.86 & 7.51 & 0.548 & 58.05 & 0.549
    \\
    
    & T2IDIff-SR &  Our-T2I-FM
    & 29.52 & 0.818 & 0.250 & 0.202 & 134.51 & 8.15 & 0.523 & 55.76 & 0.533
    \\
    \rowcolor{red!7}
    & F2IDiff-SR &  Our F2I-FM
    & \textbf{\color{red}{29.71}} & \textbf{\color{red}{0.820}} & \textbf{\color{red}{0.240}} & \textbf{\color{red}{0.190}} & \textbf{\color{red}{125.06}} & 7.37 & 0.510 & 55.74 & 0.537
    \\
    

    \hline

    \multirow{3}{4em}{RealSR (Real Datasets)} 
    & OSEDiff~\cite{OSEDiff} &  Pub-T2I-FM
    & 25.15 & 0.734 & 0.292 & 0.213 & 123.50 & \textbf{\color{red}{5.65}} & \textbf{\color{red}{0.669}} & \textbf{\color{red}{69.09}} & \textbf{\color{red}{0.634}} 
    \\
    
    & OSEDiff-38K &  Pub-T2I-FM
    & 26.53 & 0.761 & 0.242 & 0.192 & 123.13 & 6.67 & 0.497 & 62.80 & 0.592
    \\
    & T2IDiff-SR &  Our-T2I-FM
    & 26.80 & 0.765 & 0.244 & 0.194 & 113.06 & 7.17 & 0.470 & 60.52 & 0.572
    \\
    \rowcolor{red!7}
    & F2IDiff-SR &  Our-F2I-FM
    & \textbf{\color{red}{26.84}}& \textbf{\red{0.767}} & \textbf{\red{0.232}} & \textbf{\red{0.185}} & \textbf{\red{112.95}} & 6.85 & 0.462 & 60.21 & 0.581
    \\


    \hline
    \multirow{3}{4em}{DIV2K (Synthetic Dataset)} 
    & OSEDiff~\cite{OSEDiff} & Pub-T2I-FM 
    & 23.72 & 0.611 & \textbf{\color{red}{0.294}} & \textbf{\color{red}{0.198}} & \textbf{\color{red}{26.32}} & \textbf{\color{red}{4.71}} & \textbf{\color{red}{0.668}} & \textbf{\color{red}{67.97}} & \textbf{\color{red}{0.615}} 
    \\

    & OSEDiff-38K &  Pub-T2I-FM
    & 24.76 & 0.639 & 0.326 & 0.220 & 32.42 & 5.42 & 0.509 & 59.87 & 0.572
    \\
    
    & T2IDiff-SR &  Our-T2I-FM
    & 24.89 & 0.645 & 0.342 & 0.233 & 35.62 & 6.05 & 0.462 & 55.03 & 0.537
    \\    
    \rowcolor{red!7}
    & F2IDiff-SR &  Our-F2I-FM
    & \textbf{\color{red}{25.13}} & \textbf{\color{red}{0.648}} & 0.343 & 0.226 & 35.02 & 5.85 & 0.446 & 55.44 & 0.542
    \\
     
    
    \hline

  \end{tabular}
\end{table*}

\subsection{Training Datasets}
 Typically, existing T2IDiff FMs are trained on images with a maximum resolution of $2MP$, which is considerably lower than the images captured by modern smartphone cameras, often exceeding $12MP$. To develop our own diffusion-based FMs, we collected a dataset of 38,000 HR images using a flagship smartphone camera. These images were captured under optimal conditions, featuring $12MP$ resolutions and bright settings without digital zoom to ensure high-quality outputs free from blur, noise, or other artifacts. The dataset was carefully curated to ensure diversity, encompassing various locations and perspectives.

Given that our objective is not to design the most advanced generative models, we contend that large-scale training datasets are unnecessary for FMs in the context of SISR. This controlled HR training dataset for the FM enhances fidelity for the SISR task compared to public generative models, where we lack control over the training data. For the training of the T2IDiff FM, text captions were generated using pre-trained Florence models~\cite{xiao2023florence} on patches of size $512 \times 512$ extracted from the full $12MP$ images. This approach yields more concise captions compared to alternatives such as LLaVA~\cite{liu2024improved_llava}. We opted for patch-based captioning rather than full-image captioning because our SISR networks are trained on patches rather than full images, preserving higher-resolution information and ensuring practicality for smartphone camera implementation. To train the F2IDiff FM, we employed a pre-trained DINOv2-large model~\cite{oquab2023dinov2} to generate DINOv2 features on-the-fly during training. A significant advantage of using DINOv2 features over text captions on custom internal datasets lies in their fast feature generation capability during training, as opposed to the time-intensive offline text caption generation process. This approach enhances both efficiency and adaptability within our pipeline. The same dataset of 38,000 HR images was utilized for training our SISR networks. The resolution of our training images surpasses that of public datasets such as LSDIR~\cite{li2023lsdir}, which are commonly used in SISR  networks. To ensure a fair comparison with standard SOTA methods, our LR images were synthesized using the Real-ESRGAN pipeline~\cite{RealEsrgan}.

\section{Experiments}
\label{sec:experiments}
\subsection{Experimental Settings}

\textbf{Training settings:} For the development of FMs, we train three distinct U-Nets (T2IDiff, F2IDiff, Eff-F2IDiff) from scratch. All three networks share a common pre-trained encoder and decoder~\cite{Rombach_2022_CVPR_SD2.1}, initialized with pre-trained weights (f8d4). Each FM undergoes training for 30 H200 days, encompassing 700,000 iterations with a batch size of 120. Training is conducted using a learning rate of $10^{-4}$ and a patch size of $512 \times 512$, optimized with the AdamW optimizer~\cite{loshchilov2017decoupled_adamw}. We develop SISR networks, inspired by OSEDiff~\cite{OSEDiff}, based on the aforementioned FMs: T2IDiff FM, F2IDiff FM, and Eff-F2IDiff. The LoRA strategy~\cite{hu2022lora}, as outlined in OSEDiff~\cite{OSEDiff}, is applied across SISR networks, which are trained with a learning rate of $5 \times 10^{-5}$, a batch size of 6, and a patch size of $256 \times 256$, utilizing the AdamW optimizer over a period of 8 V100 days. To ensure a fair comparison with pre-trained public T2IDiff FM (SDV2.1)~\cite{Rombach_2022_CVPR_SD2.1}, we train OSEDiff~\cite{OSEDiff} on our 38K HR dataset using identical training settings to compare with our methods. This comprehensive approach underscores the efficacy and reliability of our proposed models in SISR tasks.


\begin{figure*}
  \centering
    \includegraphics[width=1\linewidth]{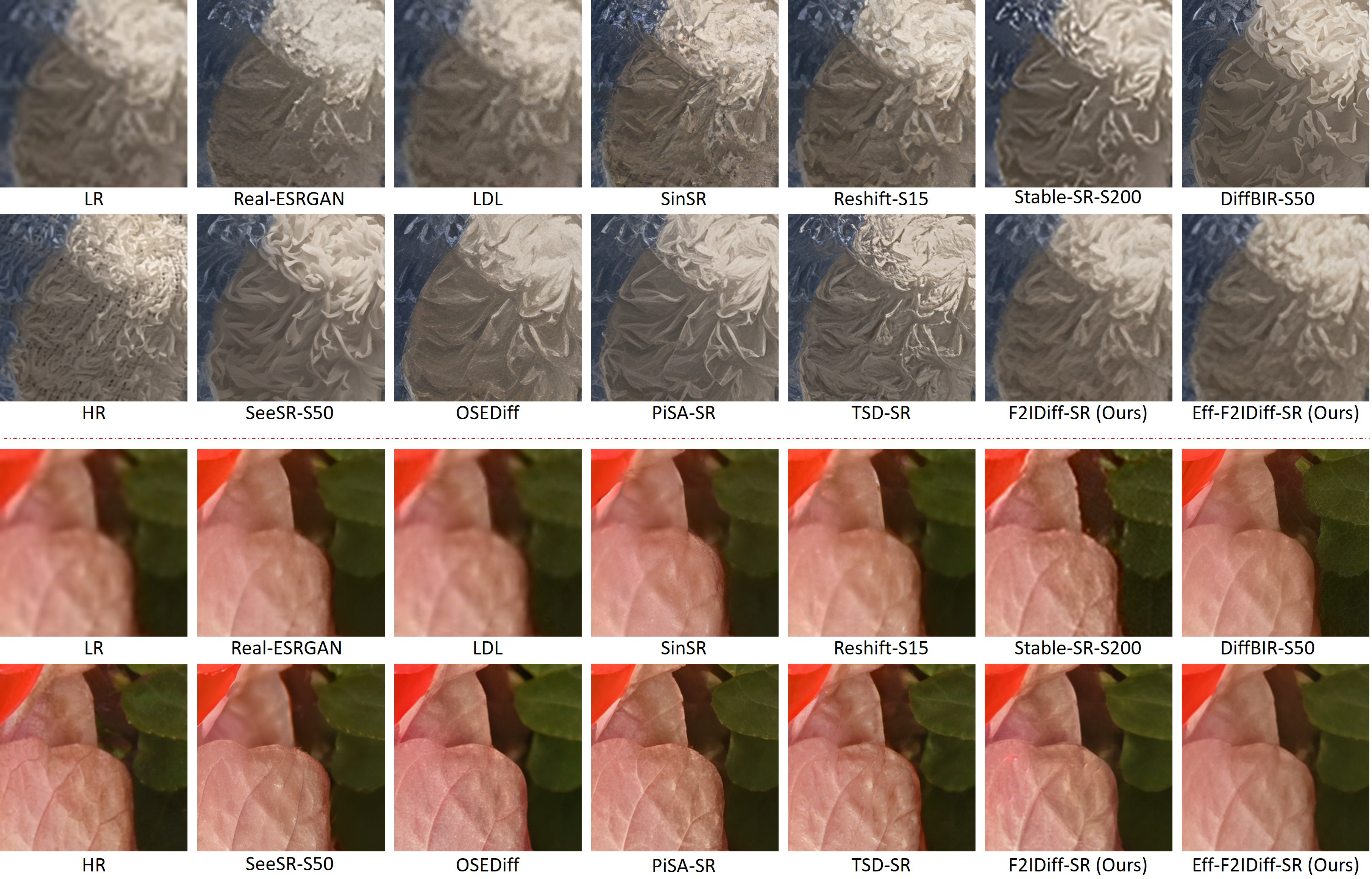}
    \caption{Zoom-in for best visuals: Qualitative comparisons of our methods (F2IDiff-SR, Eff-F2IDiff-SR) with SOTA GAN-based methods, multi-step diffusion methods, and single-step diffusion methods. The SOTA diffusion methods show significant hallucinations and unrealistic texture. Our methods generate details while preserving best fidelity.}
    \label{fig:qual_results_2}
    
\end{figure*}

\textbf{Test datasets and evaluation metrics:}
We perform comparative evaluations on standard test datasets, including DRealSR~\cite{DrealSR}, RealSR~\cite{RealSR}, and DIV2K~\cite{DIV2k}, to benchmark our results against existing methods. The synthetic test dataset consists of cropped images of size $512 \times 512$ from DIV2K, degraded using the Real-ESRGAN~\cite{RealEsrgan} degradation pipeline. For real-world test datasets, we utilize center-cropped images from RealSR and DRealSR, with LQ images resized to $128 \times 128$ and HQ images to $512 \times 512$. To comprehensively assess the performance of our SISR methods, we employ a diverse set of evaluation metrics. Fidelity metrics, including PSNR and SSIM~\cite{PSNR_SSIM}, are computed on the Y channel in the YCbCr space to measure the fidelity of SR results. Perceptual quality metrics, such as LPIPS~\cite{LPIPS} and DISTS~\cite{DISTS} (calculated in the RGB space), evaluate the perceptual quality of SR results. FID~\cite{FID} quantifies the distance between the distributions of HR and super-resolved images. Additionally, no-reference quality metrics like NIQE~\cite{NIQE}, CLIPIQA~\cite{CLIPIQA}, MUSIQ~\cite{Musiq}, and MANIQA~\cite{Maniqa} are used to evaluate image quality without reference images. This multi-faceted evaluation framework ensures a thorough analysis of our method's performance across both fidelity and perceptual quality dimensions, providing a robust assessment of its effectiveness in SISR tasks.

\subsection{Comparison of T2IDiff-SR and F2IDiff-SR}
\begin{table*}
  \caption{Quantitative comparison between F2IDiff-SR, Eff-F2IDiff-SR, and SOTA SISR methods for $4\times$ super-resolution on real-world and synthetic datasets. The best and the second-best results are highlighted in \textbf{\color{red}{red}} and \textbf{\color{blue}{blue}}. Our methods demonstrate superior performance on reference-based metrics, particularly on real-world datasets such as DRealSR~\cite{DrealSR} and RealSR~\cite{RealSR}, despite the fact that our FMs were trained on a relatively modest dataset of 38K HR images, in contrast to the Public-T2IDiff model (SDV2.1)~\cite{Rombach_2022_CVPR_SD2.1}, which was trained on billions of images. On synthetic dataset like DIV2K~\cite{DIV2k}, which are less representative of real-world scenarios encountered in smartphone cameras due to their high-resolution nature, our methods excel in fidelity metrics, including PSNR and SSIM.}
  \label{tab:quant_table_2}
  \centering
  \scriptsize
        

  \begin{tabular}{p{0.3cm}|p{2cm}|p{1.3cm}|p{0.9cm} p{0.8cm} p{0.9cm} p{1cm} p{0.7cm}|p{0.8cm} p{1.2cm} p{1cm} p{1.3cm}}
    \hline
    Data & Models & Base Method & \multicolumn{5}{c|}{Reference-based IQA - Focus on Fidelity via HR Comparison} & \multicolumn{4}{c}{Blind IQA - Focus on Perception without looking at HR} \\
    \cline{4-12}
    & & & PSNR $\uparrow$ & SSIM $\uparrow$ & LPIPS $\downarrow$ & DISTS $\downarrow$ & FID $\downarrow$ & NIQE $\downarrow$ & CLIPIQA $\uparrow$ & MUSIQ $\uparrow$ & MANIQA $\uparrow$ \\
    \hline

    \multirow{3}{4em}[-1.5ex]{\begin{turn}{90}DRealSR (Real Datasets)\end{turn}} 
    & Real-ESRGAN~\cite{RealEsrgan} & GAN  
    & 28.64 & 0.805 & 0.285 & 0.209 & 147.62 & 6.69 & 0.442 & 54.18 & 0.491
    \\
    
    & LDL~\cite{LDL_liang2022details} &  GAN
    & 28.21 & 0.813 & 0.281 & 0.213 & 155.53 & 7.13 & 0.431 & 53.85 & 0.491
    \\

    
    & Reshift-S15~\cite{Resshift} &  Diffusion
    & 28.45 & 0.763 & 0.407 & 0.270 & 175.92 & 8.28 & 0.526 & 49.86 & 0.457
    \\

    & SinSR-S1~\cite{Sinsr} &  Diffusion
    & 28.41 & 0.749 & 0.374 & 0.249 & 177.05 & 7.02 & 0.637 & 55.34 & 0.490
    \\


    & StableSR-S200~\cite{StableSR} &  Pub-T2I-FM
    & 28.04 & 0.746 & 0.335 & 0.229 & 147.03 & 6.51 & 0.617 & 58.50 & 0.560
    \\
    
    & DiffBIR-S50~\cite{Diffbir} &  Pub-T2I-FM
    & 26.84 & 0.666 & 0.445 & 0.271 & 167.38 & \textbf{\color{blue}{6.02}} & 0.629 & 60.68 & 0.590
    \\

    & SeeSR-S50~\cite{Seesr} &  Pub-T2I-FM
    & 28.26 & 0.770 & 0.320 & 0.231 & 149.86 & 6.52 & 0.667 & 64.84 & \textbf{\color{blue}{0.603}}
    \\

    & OSEDiff-S1~\cite{OSEDiff} &  Pub-T2I-FM
    & 27.92 & 0.783 & 0.297 & 0.217 & 135.29 & 6.49 & 0.696 & 64.65 & 0.590 
    \\

    & PiSA-SR-S1~\cite{PiSASR} &  Pub-T2I-FM
    & 28.31 & 0.780 & 0.296 & 0.217 & 130.61 & 6.20 & \textbf{\color{blue}{0.697}} & \textbf{\color{blue}{66.11}} & \textbf{\color{red}{0.616}}
    \\

    & TSD-SR-S1~\cite{TSDSR} &  Pub-T2I-FM
    & 27.77 & 0.756 & 0.297 & 0.214 & 134.98 & \textbf{\color{red}{5.91}} & \textbf{\color{red}{0.734}} & \textbf{\color{red}{66.62}} & 0.587
    \\

    

    \rowcolor{red!7}
    & F2IDiff-SR-S1 &  Our F2I-FM
    & \textbf{\color{red}{29.71}} & \textbf{\color{red}{0.820}} & \textbf{\color{red}{0.240}} & \textbf{\color{red}{0.190}} & \textbf{\color{red}{125.06}} & 7.37 & 0.510 & 55.74 & 0.537
    \\
    \rowcolor{red!7}
    & Eff-F2IDiff-SR-S1 &  Our F2I-FM
    & \textbf{\color{blue}{29.58}} & \textbf{\color{blue}{0.817}} & \textbf{\color{blue}{0.249}} & \textbf{\color{blue}{0.196}} & \textbf{\color{blue}{129.18}} & 7.37 & 0.483 & 56.41 & 0.532 
    \\

    \hline

    \multirow{3}{4em}[-2ex]{\begin{turn}{90}RealSR (Real Datasets)\end{turn}} 
    & Real-ESRGAN~\cite{RealEsrgan} &  GAN
    & 25.69 & 0.762 & 0.273 & 0.206 & 135.18 & 5.83 & 0.445 & 60.18 & 0.549
    \\

    & LDL~\cite{LDL_liang2022details} &  GAN
    & 25.28 & 0.757 & 0.277 & 0.212 & 142.71 & 6.00 & 0.448 & 60.82 & 0.549
    \\

    
    & Reshift-S15~\cite{Resshift} &  Diffusion
    & 26.31 & 0.741 & 0.349 & 0.250 & 142.81 & 7.27 & 0.545 & 58.10 & 0.531
    \\

    & SinSR-S1~\cite{Sinsr} &  Diffusion
    & 26.30 & 0.735 & 0.321 & 0.235 & 137.05 & 6.31 & 0.620 & 60.41 & 0.539
    \\
    
    
    & StableSR-S200~\cite{StableSR} &  Pub-T2I-FM
    & 24.69 & 0.705 & 0.309 & 0.217 & 127.20 & 5.76 & 0.619 & 65.42 & 0.621\\
    
    & DiffBIR-S50~\cite{Diffbir} &  Pub-T2I-FM
    & 24.88 & 0.667 & 0.357 & 0.229 & 124.56 & 5.63 & 0.641 & 64.66 & 0.623
    \\

    & SeeSR-S50~\cite{Seesr} &  Pub-T2I-FM
    & 25.33 & 0.727 & 0.299 & 0.221 & 125.66 & \textbf{\color{blue}{5.38}} & 0.659 & 69.37 & \textbf{\color{blue}{0.644}} 
    \\

    & OSEDiff-S1~\cite{OSEDiff} &  Pub-T2I-FM
    & 25.15 & 0.734 & 0.292 & 0.213 & 123.50 & 5.65 & 0.669 & 69.09 & 0.634 
    \\

    & PiSA-SR-S1~\cite{PiSASR} &  Pub-T2I-FM
    & 25.50 & 0.742 & 0.267 & 0.204 & 124.09 & 5.50 & \textbf{\color{blue}{0.670}} & \textbf{\color{blue}{70.15}} & \textbf{\color{red}{0.656}}
    \\

    & TSD-SR-S1~\cite{TSDSR} &  Pub-T2I-FM
    & 24.81 & 0.717 & 0.274 & 0.210 & 114.45 & \textbf{\color{red}{5.13}} & \textbf{\color{red}{0.716}} & \textbf{\color{red}{71.19}} & 0.635 
    \\

    

    
    \rowcolor{red!7}
    & F2IDiff-SR-S1 &  Our-F2I-FM
    & \textbf{\color{red}{26.84}}& \textbf{\color{red}{0.767}} & \textbf{\color{red}{0.232}} & \textbf{\color{red}{0.185}} & \textbf{\color{red}{112.95}} & 6.85 & 0.462 & 60.21 & 0.581
    \\
    
    \rowcolor{red!7}
    & Eff-F2IDiff-SR-S1 & Our-F2I-FM 
    & \textbf{\color{blue}{26.70}} & \textbf{\color{blue}{0.765}} & \textbf{\color{blue}{0.239}} & \textbf{\color{blue}{0.188}} & \textbf{\color{blue}{113.06}} & 6.90 & 0.431 & 60.43 & 0.566
    \\

    \hline
    \multirow{3}{4em}[-0.3ex]{\begin{turn}{90}DIV2K (Synthetic Datasets)\end{turn}} 
    & Real-ESRGAN~\cite{RealEsrgan} & GAN
    & 24.29 & 0.637 & 0.311 & 0.214 & 37.64 & 4.68 & 0.528 & 61.06 & 0.550
    \\
    
    & LDL~\cite{LDL_liang2022details} & GAN
    & 23.83 & 0.634 & 0.326 & 0.223 & 42.29 & 4.85 & 0.518  & 60.04  & 0.535
    \\


    & Reshift-S15~\cite{Resshift} &  Diffusion
    & 24.69 & 0.617 & 0.337 & 0.222 & 36.01 & 6.82 & 0.609 & 60.92 & 0.545
    \\

    & SinSR-S1~\cite{Sinsr} &  Diffusion
    & 24.43 & 0.601 & 0.326 & 0.207 & 35.45 & 6.02 & 0.650 & 62.80 & 0.539 
    \\


    & StableSR-S200~\cite{StableSR} &  Pub-T2I-FM
    & 23.31 & 0.573 & 0.313 & 0.214 & 24.67 & 4.76 & 0.668 & 65.63 & 0.619
    \\
    
    & DiffBIR-S50~\cite{Diffbir} &  Pub-T2I-FM
    & 23.67 & 0.565 & 0.354 & 0.213 & 30.93 & 4.71 & 0.665 & 65.66 & 0.620 
    \\

    & SeeSR-S50~\cite{Seesr} &  Pub-T2I-FM
    & 23.71 & 0.604 & 0.321 & 0.197 & 25.83 & 4.82 & 0.687 & 68.49 & \textbf{\color{blue}{0.624}}
    \\

    & OSEDiff-S1~\cite{OSEDiff} & Pub-T2I-FM 
    & 23.72 & 0.611 & 0.294 & 0.198 & 26.32 & 4.71 & 0.668 & 67.97 & 0.615 
    \\

    & PiSA-SR-S1~\cite{PiSASR} & Pub-T2I-FM
    & 23.87 & 0.606 & \textbf{\color{blue}{0.282}} & \textbf{\color{blue}{0.193}} & \textbf{\color{blue}{25.07}} & \textbf{\color{blue}{4.55}} & \textbf{\color{blue}{0.693}} & \textbf{\color{blue}{69.68}} & \textbf{\color{red}{0.640}}
    \\

    & TSD-SR-S1~\cite{TSDSR} & Pub-T2I-FM
    & 23.02 & 0.581 & \textbf{\color{red}{0.267}} & \textbf{\color{red}{0.182}} & \textbf{\color{red}{29.16}} & \textbf{\color{red}{4.32}} & \textbf{\color{red}{0.742}} & \textbf{\color{red}{71.69}} & 0.619
    \\
    
    
      
     \rowcolor{red!7}
    & F2IDiff-SR-S1 &  Our-F2I-FM
    & \textbf{\color{red}{25.13}} & \textbf{\color{red}{0.648}} & 0.343 & 0.226 & 35.02 & 5.85 & 0.446 & 55.44 & 0.542
    \\
    \rowcolor{red!7}
    & Eff-F2IDiff-SR-S1 &  Our-F2I-FM
    & \textbf{\color{blue}{25.06}} & \textbf{\color{blue}{0.645}} & 0.346 & 0.230 & 38.13 & 5.94 & 0.439 & 56.13 & 0.537
    \\
    
    \hline
    
  \end{tabular}
\end{table*}

To demonstrate that F2IDiff-SR model outperforms T2IDiff-SR model both quantitatively and qualitatively, we compare between our three trained SISR models (T2IDiff-SR, F2IDiff-SR, OSEDiff-38K) on our internal 38K HR images, along with the pre-trained OSEDiff~\cite{OSEDiff} model, as shown in Figure \ref{fig:qual_results_1}, and Table \ref{tab:quant_table_1}. OSEDiff~\cite{OSEDiff} was selected as the base pipeline because our SISR networks utilize a similar pipeline, though the underlying U-Net of diffusion-based FMs differs. To ensure a fair comparison with the public Stable Diffusion FM~\cite{Rombach_2022_CVPR_SD2.1}, we also trained OSEDiff~\cite{OSEDiff} on our internal 38K HR images and evaluated its performance against our methods. Quantitatively, our F2IDiff-SR model demonstrated superior performance on reference based metrics (PSNR, SSIM, LPIPS, DISTS, FID) on real-world test datasets (DrealSR and RealSR) as shown in Table \ref{tab:quant_table_1}.  Notably, this was achieved using only 38K training images, in contrast to the billions of images required for training the Stable Diffusion FM. On DIV2K, a synthetically generated dataset using the Real-ESRGAN degradation pipeline, our models outperformed others in terms of fidelity metrics (PSNR, SSIM). However, we believe that such extreme degradation scenarios are unlikely to occur in real-world SISR tasks due to advancements in sensor technology. This level of degradation often leads to hallucination, which is not representative of practical applications. Qualitatively, our F2IDiff-SR model demonstrates reduced hallucination and greater proximity to the HR compared to other methods, including T2IDiff-SR, OSEDiff-38K, and pre-trained OSEDiff, as illustrated in Figure \ref{fig:qual_results_1}(a). Furthermore, when applied to real-world smartphone-captured images with significantly higher resolution ($12MP$), F2IDiff exhibits consistent texture preservation and the absence of hallucination, outperforming T2IDiff-SR, OSEDiff-38K, and pre-trained OSEDiff, as depicted in Figure \ref{fig:qual_results_1}(b).

\subsection{Comparison between F2IDiff-SR, Eff-F2IDiff-SR and SOTA}
We conducted a comprehensive comparison of our proposed methods, F2IDiff-SR and Eff-F2IDiff-SR, against SOTA single-step diffusion-based SR models, including PiSA-SR~\cite{PiSASR}, OSEDiff~\cite{OSEDiff}, TSD-SR~\cite{TSDSR}, and SinSR~\cite{Sinsr}, as well as multi-step diffusion models such as Reshift~\cite{Resshift}, DiffBIR~\cite{Diffbir}, and SeeSR~\cite{Seesr}. Additionally, we evaluated our methods against GAN-based SR techniques, including Real-ESRGAN~\cite{RealEsrgan} and LDL~\cite{LDL_liang2022details}. Our analysis demonstrates that F2IDiff-SR and Eff-F2IDiff-SR outperform these methods quantitatively on reference-based metrics when applied to real-world public test datasets, such as DRealSR~\cite{DrealSR} and RealSR~\cite{RealSR}, despite the FM being trained on just 38K HR images as opposed to billion for public FM (Table \ref{tab:quant_table_2}). On the other hand, for the public DIV2K~\cite{DIV2k}, generated using synthetic degradation, our methods achieve superior performance in terms of fidelity metrics like PSNR and SSIM. However, we argue that such synthetic degradation scenarios are unlikely to occur in smartphone cameras due to the availability of HR sensors. Furthermore, we observe that increased degradation leads to greater hallucination in SISR. 

Our F2IDiff-SR method is optimized for fidelity, resulting in excellent performance on full-reference and perceptual metrics such as PSNR, SSIM, FID, LPIPS, and DISTS, outperforming all other methods by a significant margin, as illustrated in Figure \ref{fig:PSNR_SSIM_FID} and Tables \ref{tab:quant_table_1} and \ref{tab:quant_table_2}. Our method emphasizes on minimizing pixel-level reconstruction errors, preserving structural information, and aligning feature distributions closely with HR images, directly addressing hallucination issues in SISR~\cite{wang2022quality, ding2021comparison}. F2IDiff-SR sets a new SOTA in SISR, achieving SSIM of 0.820 (beating prior SOTA of 0.813), PSNR of 29.71 (surpassing prior SOTA of 28.65), and FID$\downarrow$ of 125.06 (exceeding prior SOTA of 130.61) on the DRealSR dataset. It significantly outperforms all previous methods while minimizing hallucination and preserving details.  In contrast, our methods did not achieve the best results on no-reference image quality assessment (NR-IQA) metrics, including NIQE, CLIPIQA, MUSIQ, and MANIQA, because these metrics evaluate images in isolation based on learned priors of natural scene statistics, semantic coherence, and perceptual realism—often favoring outputs with enhanced textures or stochastic details that may deviate from the original content, even when such deviations introduce inaccuracies or hallucinations ~\cite{su2025rethinking, lin2024perception}. Consequently, fidelity-oriented reconstructions that appear over-smoothed or lacking in artificial variability are penalized, despite being more faithful to the reference ~\cite{khrulkov2021neural, fang2018blind}. This unreliability of NR-IQA is illustrated in Figure \ref{fig:intro_hallucination} (first row), where OSEDiff SR result has NIQE$(\downarrow)=3.22$, CLIPIQA$(\uparrow)=0.599$,  MUSIQ$(\uparrow)=60.79$, and MANIQA$(\uparrow)=0.565$. The NR-IQA metrics for our F2IDIff method are NIQE$(\downarrow)=3.75$, CLIPIQA$(\uparrow)=0.444$,  MUSIQ$(\uparrow)=57.55$, and MANIQA$(\uparrow)=0.543$. Despite having better NR-IQA quantitative numbers for OSEDiff, it generates bird hallucination as shown in Figure \ref{fig:intro_hallucination}. Therefore, NR-IQA are not well-suited for assessing fidelity in SISR task. Qualitatively, as shown in the Figure \ref{fig:qual_results_2}, our methods exhibit the absence of hallucination and unrealistic textures, while demonstrating superior detail preservation compared to SOTA methods. In Figure \ref{fig:qual_results_2} (first example), Real-ESRGAN introduces artifacts in the towel, while LDL produces blurry outputs. Diffusion methods trained from scratch, such as SinSR and Reshift, generate unnatural textures. SOTA methods employing public T2IDiff-FM, such as Stable-SR, DiffBIR, SeeSR, OSEDiff, PiSA-SR, and TSD-SR, yield synthetic and unnatural textures. In contrast, our F2IDiff-SR and Eff-F2IDiff-SR methods achieve superior fidelity while maintaining generative capabilities.

\section{Conclusion and Future Work}
\label{sec:conclusion}
In this paper, we introduce F2IDiff-SR and Eff-F2IDiff-SR models, which are trained on DINOv2 features rather than conventional text captions. Our results demonstrate that F2IDiff-SR outperforms T2IDiff-SR on public test datasets, including DRealSR, RealSR, DIV2K, and real-world smartphone-captured images, both quantitatively and qualitatively. We also compare our methods with SOTA GAN-based methods, diffusion-based single step methods, and multi-step diffusion-based methods, showing that our methods perform best both quantitatively and qualitatively on real world test datasets (DrealSR and RealSR), despite our FM being trained on 38,000 HR images, as opposed to billions of images on which public FM was trained. On the DIV2K dataset, which is synthetically generated using the Real-ESRGAN degradation pipeline, our methods exhibit better performance in fidelity metrics such as PSNR and SSIM. However such synthetic degradation may not adequately represent the real-world scenarios encountered in smartphone cameras due to the availability of HR sensors. Future research could focus on modifying the VAE component of the FM and increasing the training dataset from 38,000 to 100,000 HR images to further enhance the controlled generative quality of the model, potentially leading to improved performance and robustness.

\newpage

{
    \small
    \bibliographystyle{ieeenat_fullname}
    \bibliography{main}
}


\end{document}